\documentclass[conference, a4paper]{IEEEtran}
\IEEEoverridecommandlockouts
\usepackage{amsmath,amssymb,amsfonts}
\usepackage{algorithmic}
\usepackage{graphicx}
\usepackage{textcomp}
\usepackage{times}
\usepackage{epsfig}
\usepackage{amsmath}
\usepackage{amssymb}
\PassOptionsToPackage{hyphens}{url}
\usepackage{adjustbox}
\usepackage{balance}
\usepackage[ruled,vlined]{algorithm2e}
\usepackage[caption=false]{subfig}

\usepackage{epsfig}
\usepackage{balance}
\def\BibTeX{{\rm B\kern-.05em{\sc i\kern-.025em b}\kern-.08em
    T\kern-.1667em\lower.7ex\hbox{E}\kern-.125emX}}

\begin{document}

\title{Fighting Deepfake by Exposing the Convolutional Traces on Images}

\author{\IEEEauthorblockN{Luca Guarnera}
\IEEEauthorblockA{\textit{Department of Mathematics}\\ \textit{and Computer Science} \\
\textit{University of Catania}\\
Catania, Italy \\
luca.guarnera@unict.it} 
\and
\IEEEauthorblockN{Oliver Giudice}
\IEEEauthorblockA{\textit{Department of Mathematics}\\ \textit{and Computer Science} \\
\textit{University of Catania}\\
Catania, Italy \\
giudice@dmi.unict.it} \\
\and
\IEEEauthorblockN{Sebastiano Battiato}
\IEEEauthorblockA{\textit{Department of Mathematics}\\ \textit{and Computer Science} \\
\textit{University of Catania}\\
Catania, Italy \\
battiato@dmi.unict.it}
}

\maketitle
\begin{abstract}
Advances in Artificial Intelligence and Image Processing are changing the way people interacts with digital images and video. Widespread mobile apps like \emph{FACEAPP} make use of the most advanced Generative Adversarial Networks (GAN) to produce extreme transformations on human face photos such gender swap, aging, etc. The results are utterly realistic and extremely easy to be exploited even for non-experienced users. This kind of media object took the name of Deepfake and raised a new challenge in the multimedia forensics field: the Deepfake detection challenge. Indeed, discriminating a Deepfake from a real image could be a difficult task even for human eyes but recent works are trying to apply the same technology used for generating images for discriminating them with preliminary good results but with many limitations: employed Convolutional Neural Networks are not so robust, demonstrate to be specific to the context and tend to extract semantics from images. In this paper, a new approach aimed to extract a Deepfake fingerprint from images is proposed. The method is based on the Expectation-Maximization algorithm trained to detect and extract a fingerprint that represents the Convolutional Traces (CT) left by GANs during image generation. The CT demonstrates to have high discriminative power achieving better results than state-of-the-art in the Deepfake detection task also proving to be robust to different attacks. Achieving an overall classification accuracy of over 98\%, considering Deepfakes from 10 different GAN architectures not only involved in images of faces, the CT demonstrates to be reliable and without any dependence on image semantic. Finally, tests carried out on Deepfakes generated by \emph{FACEAPP} achieving 93\% of accuracy in the fake detection task, demonstrated the effectiveness of the proposed technique on a real-case scenario.
\end{abstract}


\section{Introduction}
\label{sec:intro}

\begin{figure*}[t!]
    \centering
    \includegraphics[width=17cm]{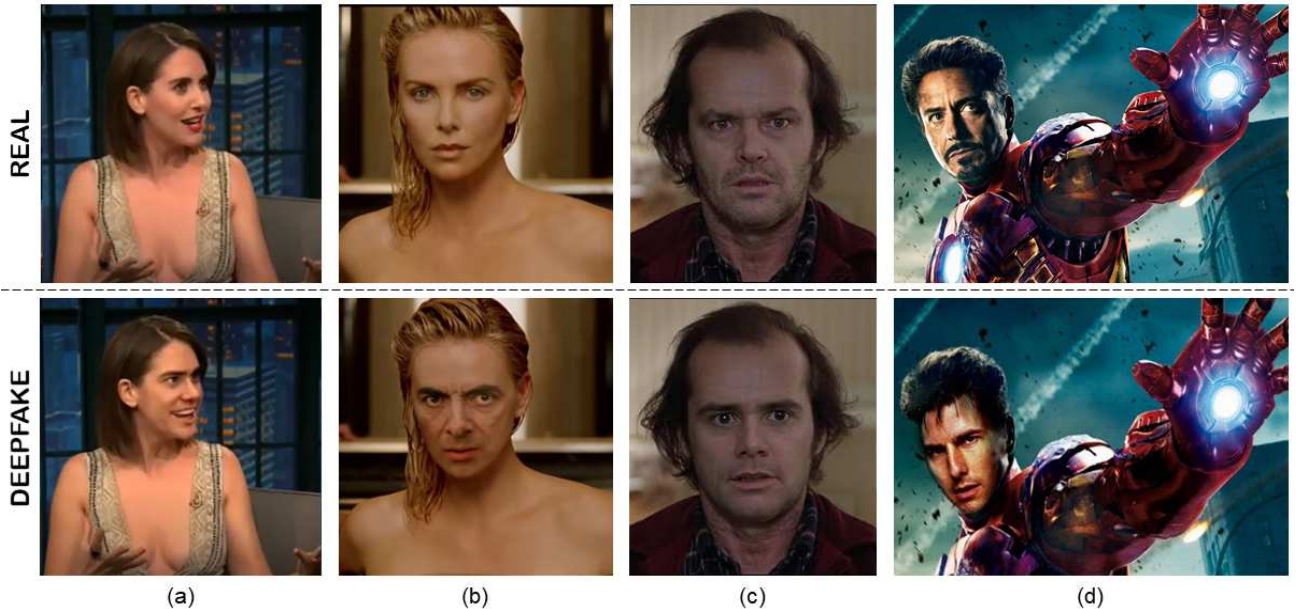}
    \caption{Examples of Deepfakes: (a) Jim Carrey's face transferred to Alison Brie's body, (b) Mr. Bean is Charlize Theron in a Deepfake version of J'adore commercial, (c) Jim Carrey instead of Jack Nicholson in Shining and (d) Tom Cruise Replaces Robert Downey Jr. in Iron Man.}
    \label{fig:1}
\end{figure*}

A digital image can be manipulated with many tools and software. Everyone with a glimpse of experience in using Photoshop or GIMP can forge photographs in order to change their contents, the semantics and - potentially - everything. However, this kind of forgery has been widely investigated throughout recent years and commercial tools with the ability to detect and describe them are also available~\cite{verdoliva2020media, piva2013overview}. The possibility to detect forgeries made with Photoshop or similar tools are related to the experience of the image manipulator being able to hide any kind of unrealistic artifact.

Advances in Artificial Intelligence, and specifically, the advent of Generative Adversarial Networks (GAN)~\cite{goodfellow2014generative}, enabled the creation and widespread of extremely refined techniques able to \emph{attack} digital data, alter it or create its contents from scratch. These tools are able to obtain surprisingly realistic results leading to the birth of the Deepfake images phenomenon, or simply Deepfakes.

In general, a Deepfake is defined as a multimedia content synthetically modified or created through automatic (or barely controlled) machine learning models. Most state-of-the-art techniques are able to do the \emph{face swap} from a source image/video to a target image/video. Recently, faces of showgirls, politicians, actors, TV presenters and many others have been the main protagonists of Deepfake attacks: one of the first example is the famous face swap of Jim Carrey on top of the the body of Alison Brie ~\footnote{\url{https://www.youtube.com/watch?v=SEar_6UtX9U}} (Figure~\ref{fig:1}a), or Mr. Bean and Charlize Theron in the Deepfake version of the commercial of J'adore~\footnote{\url{https://www.youtube.com/watch?v=gZVdPJhBkqg}} (Figure~\ref{fig:1}b),  and again Jim Carrey instead of Jack Nicholson in Shining~\footnote{\url{https://www.youtube.com/watch?v=JbzVhzNaTdI}} (Figure~\ref{fig:1}c), or Tom Cruise replacing Robert Downey Jr. in Iron Man~\footnote{\url{https://www.youtube.com/watch?v=iDM69UEyM3w}} (Figure~\ref{fig:1}d). 

Deepfakes are not only involved in face-related tasks but they could be engaged to swap or generate realistic places, animals, object, etc. Indeed, this could  bring disruptive innovation in many working areas, such as in the automotive industry or in architecture, since it is possible to generate a car or an apartment through dedicated GANs or in the film industry where it is possible, when necessary, to replace the face of a stuntman with an actor; but, on the other hand it could lead to serious social repercussions, privacy issues and major security concerns. For example, there are many Deepfake videos connected to the world of porn used to discredit famous actresses like Emma Watson o Angelina Jolie, or they can be used to spread disinformation and fake news. Moreover, the creation of Deepfakes is becoming extremely easy: widespread mobile apps like \emph{FACEAPP}~\footnote{\url{https://www.faceapp.com/}}, are able to produce transformations on human faces such gender swap, aging, etc. The results are utterly realistic and extremely easy to produce even for non-experienced users with a few taps on their mobile phone.

It is clear that the Deepfake phenomenon raises a serious safety issue and it is absolutely necessary to create new techniques able to detect and counteract it \cite{verdoliva2020media, guarnera2020preliminary}.

%Among the big companies that have decided to fight the deepkake phenomenon there are Google and Facebook. The first one, to meet the needs of researchers in this field, created a database of fake videos ~\cite{rossler2019faceforensics++} while the second, together with Microsoft, launched the Deepfake Detection Challenge initiative ~\footnote{\url{https://deepfakedetectionchallenge.ai/}}.

While detecting a Deepfake is difficult for humans, recent works have shown that they could be detected surprisingly easy by employing Convolutional Neural Networks (CNN) specifically trained on the task. However, CNN solutions presented till today, lack of robustness, generalizing and explainability. They are extremely specific to the context in which they were trained and, being very deep, tend to extract the underlying semantics from images without inferring any unique fingerprint. A detailed discussion about such limits will be dealt with in the final part of the paper.

In order to find a unique fingerprint related to the specific GAN architecture that created the Deepfake image, in this paper an extension of our previous work \cite{guarnera2020deepfake} is presented. The fingerprint extraction method based on the Expectation-Maximization Algorithm will be furtherly discussed focusing on its capabilities to extract the Convolutional Traces (CT) embedded by the generative process. The CT could be employed in many related classification tasks but in this paper the finalized pipeline for fakeness detection is finalized with the adoption of a Random Forest classifier. Moreover, the method was deeply tested for robustness with many  attacks carried out on images before the extraction of the CT. Also generalizing was demonstrated by testing real images against images generated by ten different GAN architectures, which is the widest test carried out on the task till today. Comparison with state-of-the-art methods demonstrated that the overall approach achieves in almost all cases best classification results. Moreover, we would like to highlight that different state-of-the-art methods for Deepfake detection used approaches based on CNN and these require extremely computational demanding (both for hardware and for time needed), while the proposed approach achieves excellent classification results using only the CPU power of a common laptop.

The remainder of this paper is organized as follows: Section~\ref{sec:related} presents state-of-the-art Deepfake generation and detection methods. The proposed approach to extract the Convolutional Trace is described in Section \ref{sec:approach}. Classification phase and experimental results are reported in Section~\ref{sec:results}. In Section~\ref{sec:RobustTest} the proposed approach is  demonstrated to be robust to different attacks. Finally, obtained classification results were compared with recent state-of-the-art methods in Section~\ref{sec:CompDetMeth}. Section~\ref{sec:conclusion} concludes the paper.

\section{Related Works}
\label{sec:related}

\begin{figure}[t!]
    \centering
    \includegraphics[width=8cm]{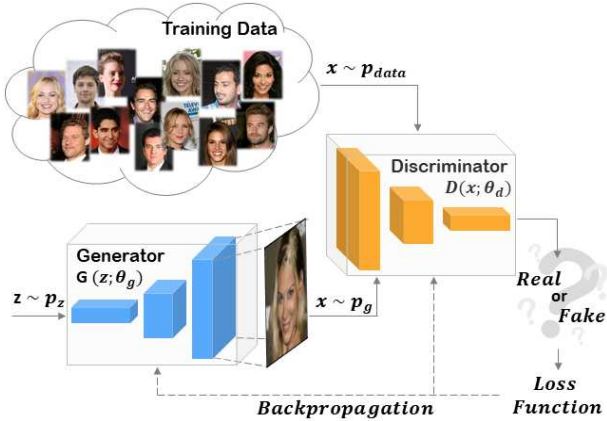}
    \caption{Schematic description of a GAN learning framework.}
    \label{fig:genericalGAN}
\end{figure}

Deepfakes are generally created by techniques based on Generative Adversarial Networks (GANs) firstly introduced by Goodfellow et al.~\cite{goodfellow2014generative}. In \cite{goodfellow2014generative}, authors proposed a new framework in which two models simultaneously train: a generative model $G$, that captures the data distribution, and a discriminative model $D$, able to estimate the probability that a sample comes from the training data rather than from $G$. The training procedure for $G$ is to maximize the probability of $D$ making a mistake thus resulting in a min-max two-player game.
Mathematically, the generator accepts a random input $z$ with density $p_z$ and returns an output $x = G (z, \Theta_g)$ according to a certain probability distribution $p_g$ ($\Theta_g$ represents the parameters of the generative model). The discriminator, $D(x, \Theta_d)$ computes the probability that $x$ comes from the distribution of training data $p_{data}$ ($\Theta_d$ represents the parameters of the discriminative model). The overall objective is to obtain a generator, after the training phase, which is a good estimator of $p_{data}$. When this happens, the discriminator is ``deceived" and will no longer be able to distinguish the samples from $p_{data}$ and $p_g$; therefore $p_g$ will follow the targeted probability distribution, i.e. $p_{data}$. Figure~\ref{fig:genericalGAN} shows a simplified description of a GAN framework.
In the case of Deepfakes, $G$ can be thought as a team of counterfeiters trying to produce fake currency, while $D$ stands to the police, trying to detect the malicious activity. $G$ and $D$ can be implemented as any kind of generative model, in particular when deep neural networks are employed results become extremely accurate.
Through recent years, many GAN architectures were proposed for different applications e.g., image to image translation~\cite{zhu2017unpaired}, image super resolution~\cite{ledig2017photo}, image completion~\cite{iizuka2017globally}, and text-to-image generation~\cite{reed2016generative}.

\subsection{Deepfake Generation Techniques for Faces}
Advances in GAN architectures lead to different works  dealing with human faces. STARGAN, created by Choi et al.~\cite{choi2018stargan}, is a method capable of performing image-to-image translations on multiple domains using a single model (e.g, change hair color, facial expression). Many methods works in the latent space representation in order to set constraints to the attributes to be modified, an example is ATTGAN, created by He et al.~\cite{he2019attgan}. Cho et al.~\cite{cho2019image} proposed the "group-wise deep whitening-and coloring method" (GDWCT) for a better styling capacity, obtaining a great improvement in the image translation and style transfer task in terms of computational efficiency and quality of generated images. The stage changes when surprising results of Deepfake images were obtained by Style Generative Adversarial Network (STYLEGAN)~\cite{karras2019style}. STYLEGAN was used to create the so-called "this person does not exist" website~\footnote{\url{https://thispersondoesnotexist.com/}}. Moreover, a few imperfect artifact created by STYLEGAN were fixed by Karras et al.~\cite{karras2020analyzing} with improvements to the generator (including re-designed normalization, multi-resolution, and regularization methods), creating the even more realistic images with the so called STYLEGAN2.

\subsection{Deepfake detection methods}
\label{sec:detection_methods}
A starting point to detect Deepfakes is indeed the analysis in the Fourier domain which is a well known technique to find anomalies for image forensics experts \cite{battiato2016multimedia}. Indeed, some Deepfake images, in the Fourier domain, after being processed by a Discrete Fast Fourier Transform, show abnormal frequencies distributions. This preliminary insight was detected by Guarnera et al. \cite{guarnera2020preliminary} in which the authors tried preliminarly to detect Deepfakes by means of well-known forgery detection tools (\cite{piva2013overview,battiato2016multimedia,giudice20191}) with only few insights for future works as results. The analysis in the Fourier domain was employed more in deep by Zhang et al.~\cite{zhang2019detecting} in a naive strategy achieved good performances. Later, an interesting work known as FakeSpotter was proposed by Wang et al.~\cite{wang2019fakespotter}. They described a new method based on monitoring neuron behaviors of a dedicated CNN to detect faces generated by Deepfake technologies. The comparison with Zhang et al.~\cite{zhang2019detecting} demonstrated an average detection accuracy of more than 90\% %w.r.t. 65.7\% of Zhang et al.~\cite{zhang2019detecting}.

Wang et al.~\cite{Wang_2020_CVPR} trained a ResNet-50 to discriminate real images from those generated by ProGAN~\cite{karras2017progressive} and demonstrated that the trained model is able to generalize for the detection of Deepfakes generated by other architectures than ProGAN. They also demonstrate to achieve good robustness to JPEG compression, spatial blurring and scaling transformations.

Jain et al.~\cite{jain2020detecting} proposed a work known as DAD-HCNN, a new framework based on a hierarchical classification pipeline composed of three levels to distinguish respectively real Vs altered images (first level), retouched Vs GAN’s generated images (second level) and finally, the specific GAN architecture (third level). From the conducted tests, the framework can detect retouching and GANs generated images with high accuracy.

A reference dataset was introduced by Rossler et al.~\cite{rossler2019faceforensics++} as a benchmark for fake detection. It is called FaceForensics++, and is based mainly on four manipulation methods: two computer graphics-based methods (Face2Face~\cite{thies2016face2face}, FaceSwap~\footnote{\url{https://github.com/MarekKowalski/FaceSwap/}}) and 2 learning-based approaches (DeepFakes~\footnote{\url{https://github.com/deepfakes/faceswap/}}, NeuralTextures~\cite{thies2019deferred}). 

%They addressed the problem of fake detection as a binary classification problem for each frame of manipulated videos, considering different techniques present in the state of the art \cite{afchar2018mesonet, bayar2016deep, chollet2017xception, cozzolino2017recasting, fridrich2012rich, rahmouni2017distinguishing}. 

\begin{figure*}[]
    \centering
    \includegraphics[width=1\linewidth]{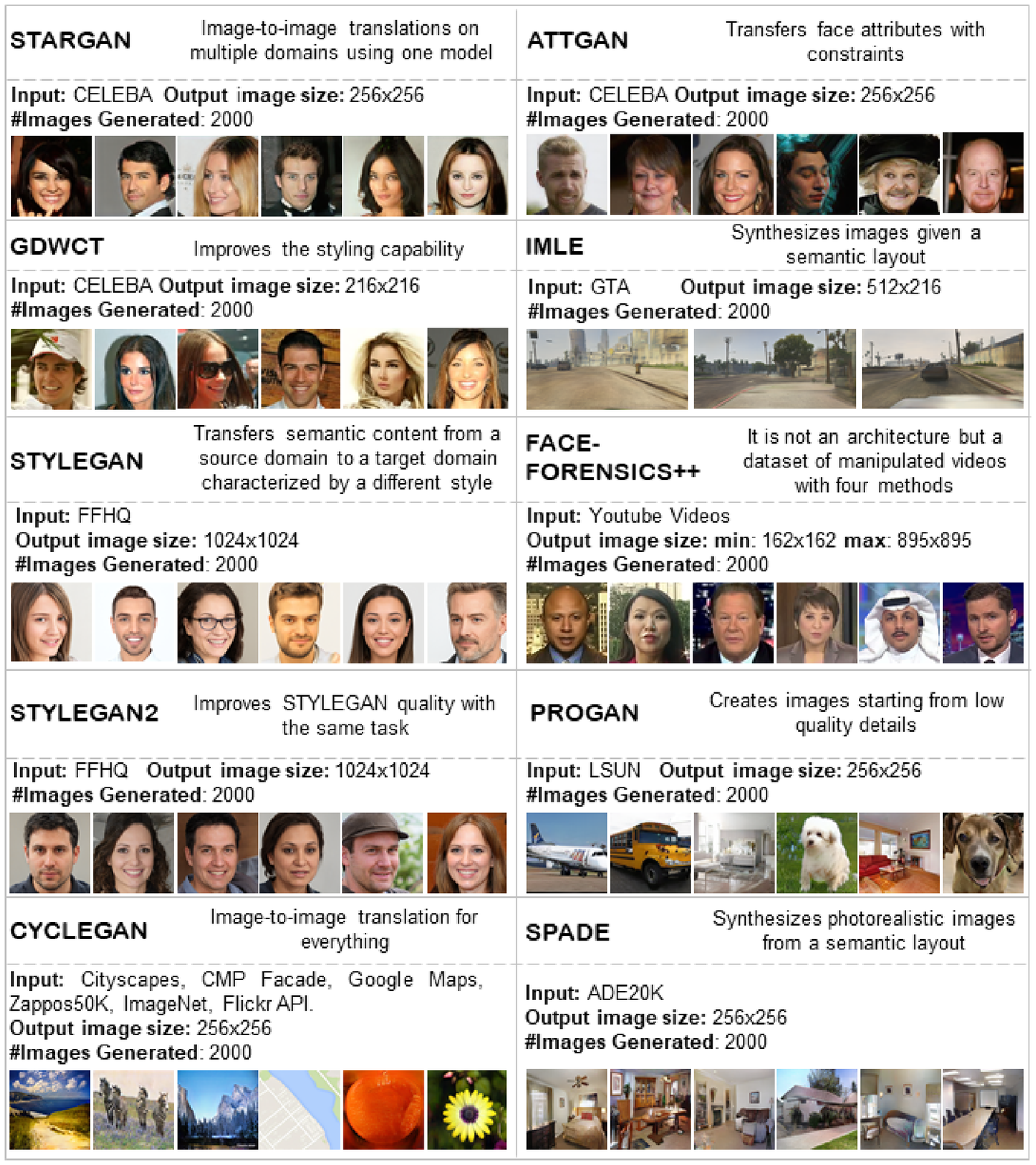}
    \caption{Details for each image set used in this paper. On the right of each deep architecture's name is reported a brief description. \textit{Input} represents the dataset used for both training and test phase of the respective architecture. \textit{Image size} describes the image size of the generated Deepfakes dataset. As regards FACEFORENSICS++ is concerned that for each video frame, the patch referring to the face, is detected and extracted automatically. This patch could have different sizes. \textit{\#Images Generated} describes the total number of images taken into account for the considered architecture. Finally, image examples are reported.}
    \label{fig:NetworksDetail}
\end{figure*}

By roughly considering literature in the field, it seems like that Deepfake detection is an easy task, already solved. However, analytical techniques based on frequency domain still lack of accuracy and CNN techniques while achieving good results tend to discriminate semantics more than GAN-specific traces. Moreover CNN techniques are computationally intensive and difficult to be understood or controlled \cite{hulzebosch2020detecting}. 
To overcome this, Guarnera et al.~\cite{guarnera2020deepfake} proposed a new analytical solution to extract an unique fingerprint from images that was demonstrated to be specific to the GAN that generated the image itself. In this paper, the technique will be presented in the mathematical details with furtherly discussion on robustness and generalization, by means of the many carried out experiments: the widest test cases, as today, in the Deepfake detection task will be presented. For this purpose we employed ten of the most famous and effective Deepfake generation architectures: CYCLEGAN~\cite{zhu2017unpaired}, STARGAN~\cite{choi2018stargan}, ATTGAN~\cite{he2019attgan}, GDWCT~\cite{cho2019image},
STYLEGAN~\cite{karras2019style}, STYLEGAN2~\cite{karras2020analyzing}, PROGAN~\cite{karras2017progressive}, FACEFORENSICS++~\cite{rossler2019faceforensics++}, IMLE~\cite{li2019diverse} and SPADE~\cite{park2019semantic}. Figure~\ref{fig:NetworksDetail} resumes the differences of these techniques in terms of image size, datasets used as input, goal and examples of generated images. For each architecture 2000 images were generated.

\section{Extracting Convolutional Traces}
\label{sec:approach}

\begin{figure*}[t!]
\begin{center}
\includegraphics[width=1\linewidth]{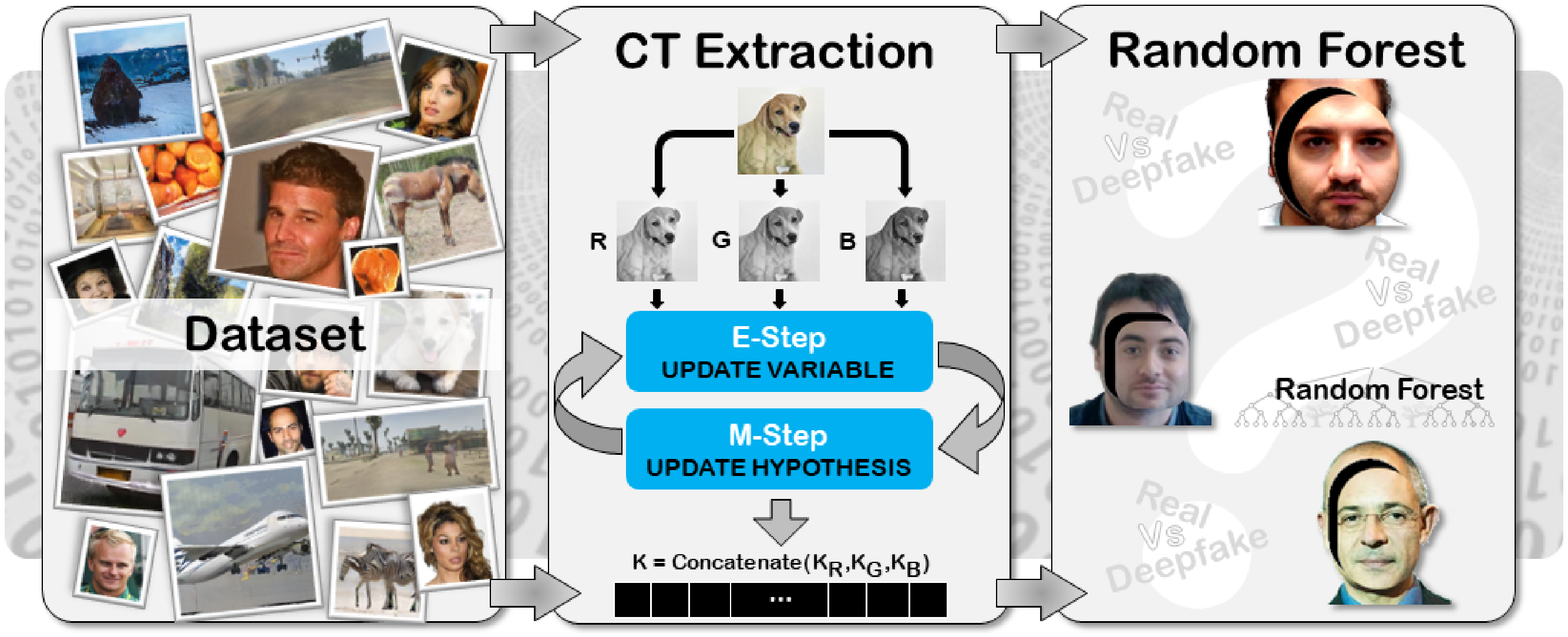}
\end{center}
   \caption{Overall finalized Deepfake detection pipeline. The dataset block represents an overview of input data used in this work (Real and Deepfake images).
   For each image we apply EM algorithm on every channel (R,G,B) obtaining $K_R$, $k_B$, $K_G$ feature vectors; the concatenation of them gives the final representation ($K$) of the input image: the so called Convolutional Trace (CT). Finally, the CT is employed to discriminate real from Deepfake images by means of Random Forest.}
\label{fig:Pipeline}
\end{figure*}

Generative Adversarial Networks (GAN) are used to generate Deepfakes. Once trained, the fundamental element involved in the image creation is the generator $G$ which is composed of Transpose Convolution layers~\cite{radford2015unsupervised}. They apply kernels to the input image, similarly to kernels in Convolutional Layer but they act inversely in order to obtain an output larger but proportional to the input dimensions. Thus, the image creation pipeline is different from the pipeline commonly used in a camera device  in which each step introduces typical noise that is then used for naive image forgery detection \cite{battiato2016multimedia}. However, the image creation process related to the Transpose Convolution layers of GAN should be consistent and identifiable in local correlations of pixels in the spatial RGB space. To find these traces, an Expectation-Maximization (EM) algorithm~\cite{moon1996expectation} was employed in order to define a conceptual mathematical model able to capture the pixel correlation in the images (e.g. spatially) and discriminate between two distributions: the expected one (natural) and others (possibly Deepfake). The result of EM is a feature vector representing the structure of the Transpose Convolution Layers employed during the generation of the image, encoding in some sense if such image is a Deepfake or not, thus it can be called \emph{Convolutional Trace} (CT).

The CT extraction techniques works as follows. The initial goal is to extract a description, from input image $I$, able to numerically represent the local correlations between each pixel in a neighbourhood. This can be done by means of convolution with a kernel $k$ of $N \times N$ size: 

\begin{equation}
	\label{eq:eqconv}
	I[x,y] = \sum\limits_{s,t=-\alpha}^\alpha k_{s,t}*I[x+s,y+t] 
\end{equation}

In Equation~\ref{eq:eqconv}, the value of the pixel $I[x,y]$ is computed considering a neighborhood of size $N \times N$ of the input data. It is clear that the new estimated information $I[x,y]$ mainly depends on the kernel used in the convolution operation, which establishes a mathematical relationship between the pixels. For this reason, our goal is to define a vector $k$ of size $N \times N$ able to capture this hidden and implicit relationship which characterizes the forensic trace we want to exploit. 

Let's assume that the element $I[x,y]$ belongs to one of the following models:
\begin{itemize}
    \item $M_{1}$: when the element $I[x,y]$ satisfies Equation~\ref{eq:eqconv};
    \item $M_{2}$: otherwise.
\end{itemize}

The EM algorithm is employed with its two different steps:
\begin{enumerate}
    \item \textbf{Expectation step}: computes the (density of) probability that each element belongs to model ($M_{1}$ or $M_{2}$);
    \item \textbf{Maximization step}: estimates the (weighted) parameters based on the probabilities of belonging to instances of ($M_{1}$ or $M_{2}$).
\end{enumerate}

Let's suppose that $M_{1}$ and $M_{2}$ have different probability distributions with $M_{1}$  Gaussian distribution with zero mean and unknown variance  and $M_{2}$ uniform. In the Expectation step, the Bayes rule that $I[x, y]$ belongs to the model $M_{1}$ is computed as follows:

\begin{equation}
	\label{eq:Bayes}
	%\color{red}
	\begin{split}
	    Pr\{I[x,y] \in M_{1} \mid I[x,y]\} = \\
		= \frac{Pr\{I[x,y] \mid I[x,y] \in M_{1}\}*Pr\{I[x,y] \in M_{1}\}}{\sum\limits_{i=1}^2{Pr\{I[x,y] \mid I[x,y] \in M_{i}\}*Pr\{I[x,y] \in M_{i}\}}}
	\end{split}
\end{equation}

where the probability distribution of $M_{1}$ which represents the probability of observing a sample $I[x,y]$, knowing that it was generated by the model $M_{1}$ is:

\begin{equation}
	\label{eq:Bayes2}
	%\color{red}
	\begin{split}
	Pr\{I[x,y] \mid I[x,y] \in M_{1}\} =
	\frac{1}{\sigma\sqrt{2\pi}}e^{-\frac{(R[x,y])^2}{2\sigma^2}}
	\end{split}
\end{equation}
where
\begin{equation}
	\label{eq:BayesX}
	%\color{red}
	\begin{split}
	R[x,y]=\bigg|I[x,y]-\sum\limits_{s,t=-\alpha}^\alpha{k_{s,t}I[x+s,y+t]}\bigg|
	\end{split}
\end{equation}.

The variance value $\sigma^2$, which is still unknown, is then estimated in the Maximization step. Once defined if $I[x,y]$ belongs to model $M_{1}$ (or $M_{2}$), the values of the vector $\vec k$ are estimated using Least Squares method, minimizing the following:

\begin{equation}
	\label{eq:minimiz}
	%\color{red}
	E(\vec k) = \sum\limits_{x,y}w[x,y]\Bigg(I[x,y]-\sum\limits_{s,t=-\alpha}^\alpha{k_{s,t}I[x+s,y+t]} \Bigg)^2
\end{equation}

where $w \equiv Pr\{I[x,y] \in M_{1} \mid I[x,y]\}$ (\ref{eq:Bayes}). This error function (\ref{eq:minimiz}) is minimized by computing the gradient of vector $\vec k$. The update of $k_{i, j}$ is carried out by computing the partial derivative of~(\ref{eq:minimiz}) as follows:

\begin{equation}
	\label{eq:derivata0}
	%\color{red}
	\frac{\partial E}{\partial k_{i,j}} = 0
\end{equation}

Hence, the following linear equations system is obtained:

\begin{equation}
	\label{eq:derivata}
	%\color{red}
	\begin{split}
	\sum\limits_{s,t=-\alpha}^\alpha k_{s,t}\Bigg(\sum\limits_{x,y}w[x,y]I[x+i, y+j]I[x+s, y+t]\Bigg) = \\       
	=\sum\limits_{x,y}w[x,y]I[x+i, y+j]I[x, y]
	\end{split}
\end{equation}

The two steps of the EM algorithm are iteratively repeated. The algorithm is applied to each channel of the input image (RGB color space). 

The obtained feature vector $\vec k$ is the desired CT and has dimensions dependent on parameter $\alpha$. Note that the element $k_{0,0}$ will always be set equal to 0 ($k_{0,0}=0$). Thus, for example, if a kernel $k$ with $3 \times 3$ size is employed, the resulting $\vec k$ will be a vector of $24$ elements (since the values $k_{0,0}$ are excluded). This is obtained by concatenating the features extracted from each of the three RGB channels.

The computational complexity of the EM algorithm can be estimated to be linear in $d$ (the number of characteristics of the input data taken into consideration), $n$ (the number of objects) and $t$ (the number of iterations) making it easily to be computed in seconds on a common laptop.

Two aspects are of extreme importance: (i) the proposed CT extraction technique does not need training, it is applied on images and extracts a discriminative feature vector; (ii) the CT extraction is not a deep learning architecture, thus it is not able to encode high level information such semantics.  This will be demonstrated in the following Sections with experimental tests.

\section{Classification of Deepfakes}
\label{sec:results}

In this Section, the Convolutional Trace (CT) extracted by means of the technique presented in Section \ref{sec:approach}, will be demonstrated to have great discriminative power for the Deepfake detection task. Moreover, the independence on image semantics will be demonstrated in this Section by testing against Deepfakes not representing merely faces.

\begin{table*}[t!]
\centering 
\caption{Overall accuracy between CELEBA vs. each of the considered GAN. Results are presented w.r.t. all the different kernel sizes ($3x3$, $5x5$, $7x7$) and with different classifiers: KNN, with $k \in \{3,5,7,9,11,13\}$; Linear SVM, Linear Discriminant Analysis (LDA).}
\begin{tabular}{ccllcccccccccccc}
\cline{2-16}
\multicolumn{1}{l|}{}                &

\multicolumn{3}{c|}{\textbf{ATTGAN}} & \multicolumn{3}{c|}{\textbf{CYCLEGAN}} & \multicolumn{3}{c|}{\textbf{FACEFORENSICS++}} & \multicolumn{3}{c|}{\textbf{GDWCT}}  & \multicolumn{3}{c|}{\textbf{IMLE}}      \\ \cline{2-16} 
\multicolumn{1}{l|}{}                & \multicolumn{3}{c|}{\textbf{Kernel Size}}                                                    & \multicolumn{3}{c|}{\textbf{Kernel Size}}                                                      & \multicolumn{3}{c|}{\textbf{Kernel Size}}                                                           & \multicolumn{3}{c|}{\textbf{Kernel Size}}                                                     & \multicolumn{3}{c|}{\textbf{Kernel Size}}                                                       \\
\multicolumn{1}{l|}{}                & \textbf{3x3}          & \multicolumn{1}{c}{\textbf{5x5}} & \multicolumn{1}{c|}{\textbf{7x7}} & \textbf{3x3}              & \textbf{5x5}             & \multicolumn{1}{c|}{\textbf{7x7}}       & \textbf{3x3}               & \textbf{5x5}               & \multicolumn{1}{c|}{\textbf{7x7}}         & \textbf{3x3}              & \textbf{5x5}             & \multicolumn{1}{c|}{\textbf{7x7}}      & \textbf{3x3}              & \textbf{5x5}              & \multicolumn{1}{c|}{\textbf{7x7}}       \\ \hline
\multicolumn{1}{|c|}{\textbf{3-NN}}  & \textbf{92.62}        & 82.92                            & \multicolumn{1}{l|}{81.51}        & \textbf{93.59}            & 86.74                    & \multicolumn{1}{c|}{87.29}              & \textbf{97.31}             & 84.74                      & \multicolumn{1}{c|}{80.70}                & \textbf{91.38}            & 72.19                    & \multicolumn{1}{c|}{72.49}             & \textbf{97.76}            & 97.19                     & \multicolumn{1}{c|}{94.73}              \\
\multicolumn{1}{|c|}{\textbf{5-NN}}  & \textbf{92.99}        & 84.47                            & \multicolumn{1}{l|}{80.21}        & \textbf{93.33}            & 86.64                    & \multicolumn{1}{c|}{87.63}              & \textbf{96.98}             & 84.21                      & \multicolumn{1}{c|}{78.57}                & \textbf{91.08}            & 75.03                    & \multicolumn{1}{c|}{73.54}             & \textbf{97.69}            & 96.93                     & \multicolumn{1}{c|}{94.64}              \\
\multicolumn{1}{|c|}{\textbf{7-NN}}  & \textbf{92.99}        & 85.20                            & \multicolumn{1}{l|}{80.47}        & \textbf{93.25}            & 85.85                    & \multicolumn{1}{c|}{86.50}              & \textbf{96.56}             & 83.68                      & \multicolumn{1}{c|}{79.07}                & \textbf{91.08}            & 75.60                    & \multicolumn{1}{c|}{75.53}             & \textbf{97.39}            & 96.85                     & \multicolumn{1}{c|}{94.64}              \\
\multicolumn{1}{|c|}{\textbf{9-NN}}  & \textbf{92.71}        & 84.68                            & \multicolumn{1}{l|}{80.47}        & \textbf{93.25}            & 85.66                    & \multicolumn{1}{c|}{86.84}              & \textbf{96.56}             & 82.95                      & \multicolumn{1}{c|}{79.07}                & \textbf{91.58}            & 75.71                    & \multicolumn{1}{c|}{75.00}             & \textbf{97.24}            & 96.59                     & \multicolumn{1}{c|}{94.25}              \\
\multicolumn{1}{|c|}{\textbf{11-NN}} & \textbf{92.90}        & 84.68                            & \multicolumn{1}{l|}{79.56}        & \textbf{93.00}            & 85.07                    & \multicolumn{1}{c|}{86.50}              & \textbf{96.47}             & 81.89                      & \multicolumn{1}{c|}{79.07}                & \textbf{91.48}            & 75.48                    & \multicolumn{1}{c|}{74.34}             & \textbf{96.94}            & 96.50                     & \multicolumn{1}{c|}{94.06}              \\
\multicolumn{1}{|c|}{\textbf{13-NN}} & \textbf{92.52}        & 84.99                            & \multicolumn{1}{l|}{78.91}        & \textbf{92.32}            & 84.97                    & \multicolumn{1}{c|}{85.94}              & \textbf{96.31}             & 81.79                      & \multicolumn{1}{c|}{79.07}                & \textbf{91.28}            & 75.94                    & \multicolumn{1}{c|}{73.68}             & 96.42                     & \textbf{96.59}            & \multicolumn{1}{c|}{94.06}              \\
\multicolumn{1}{|c|}{\textbf{SVM}}   & \textbf{90.19}        & 88.51                            & \multicolumn{1}{l|}{87.11}        & 78.90                     & 83.20                    & \multicolumn{1}{c|}{\textbf{90.33}}     & \textbf{92.70}             & 84.74                      & \multicolumn{1}{c|}{83.33}                & \textbf{89.30}            & 77.30                    & \multicolumn{1}{c|}{80.95}             & 95.67                     & \textbf{97.27}            & \multicolumn{1}{c|}{95.40}              \\
\multicolumn{1}{|c|}{\textbf{LDA}}   & \textbf{90.47}        & 87.47                            & \multicolumn{1}{l|}{86.59}        & 77.05                     & 83.30                    & \multicolumn{1}{c|}{\textbf{89.65}}     & \textbf{92.28}             & 84.53                      & \multicolumn{1}{c|}{81.70}                & \textbf{88.90}            & 77.41                    & \multicolumn{1}{c|}{82.01}             & 95.00                     & \textbf{96.59}            & \multicolumn{1}{c|}{94.64}              \\ \hline
\multicolumn{1}{l}{}                 & \multicolumn{1}{l}{}  &                                  &                                   & \multicolumn{1}{l}{}      & \multicolumn{1}{l}{}     & \multicolumn{1}{l}{}                    & \multicolumn{1}{l}{}       & \multicolumn{1}{l}{}       & \multicolumn{1}{l}{}                      & \multicolumn{1}{l}{}      & \multicolumn{1}{l}{}     & \multicolumn{1}{l}{}                   & \multicolumn{1}{l}{}      & \multicolumn{1}{l}{}      & \multicolumn{1}{l}{}                    \\ \cline{2-16} 
\multicolumn{1}{l|}{}                & \multicolumn{3}{c|}{\textbf{PROGAN}} & \multicolumn{3}{c|}{ \textbf{SPADE}}    & \multicolumn{3}{c|}{\textbf{STARGAN}}        & \multicolumn{3}{c|}{ \textbf{STYLEGAN}} & \multicolumn{3}{c|}{\textbf{STYLEGAN2}} \\ \cline{2-16} 
\multicolumn{1}{l|}{}                & \multicolumn{3}{c|}{\textbf{Kernel Size}}                                                    & \multicolumn{3}{c|}{\textbf{Kernel Size}}                                                      & \multicolumn{3}{c|}{\textbf{Kernel Size}}                                                           & \multicolumn{3}{c|}{\textbf{Kernel Size}}                                                     & \multicolumn{3}{c|}{\textbf{Kernel Size}}                                                       \\
\multicolumn{1}{l|}{}                & \textbf{3x3}          & \multicolumn{1}{c}{\textbf{5x5}} & \multicolumn{1}{c|}{\textbf{7x7}} & \textbf{3x3}              & \textbf{5x5}             & \multicolumn{1}{c|}{\textbf{7x7}}       & \textbf{3x3}               & \textbf{5x5}               & \multicolumn{1}{c|}{\textbf{7x7}}         & \textbf{3x3}              & \textbf{5x5}             & \multicolumn{1}{c|}{\textbf{7x7}}      & \textbf{3x3}              & \textbf{5x5}              & \multicolumn{1}{c|}{\textbf{7x7}}       \\ \hline
\multicolumn{1}{|c|}{\textbf{3-NN}}  & \textbf{95.70}        & 83.38                            & \multicolumn{1}{l|}{78.76}        & \textbf{96.72}            & 78.35                    & \multicolumn{1}{c|}{84.96}              & \textbf{88.40}             & 82.08                      & \multicolumn{1}{c|}{83.73}                & 94.62                     & \textbf{99.48}           & \multicolumn{1}{c|}{98.95}             & 96.58                     & 98.06                     & \multicolumn{1}{c|}{\textbf{98.82}}     \\
\multicolumn{1}{|c|}{\textbf{5-NN}}  & \textbf{95.85}        & 82.24                            & \multicolumn{1}{l|}{81.08}        & \textbf{96.64}            & 78.35                    & \multicolumn{1}{c|}{85.15}              & \textbf{88.10}             & 82.31                      & \multicolumn{1}{c|}{83.46}                & 95.29                     & \textbf{99.35}           & \multicolumn{1}{c|}{99.12}             & 96.91                     & 98.06                     & \multicolumn{1}{c|}{\textbf{99.15}}     \\
\multicolumn{1}{|c|}{\textbf{7-NN}}  & \textbf{95.54}        & 83.86                            & \multicolumn{1}{l|}{82.08}        & \textbf{96.27}            & 79.20                    & \multicolumn{1}{c|}{85.15}              & \textbf{87.88}             & 81.63                      & \multicolumn{1}{c|}{82.41}                & 94.72                     & \textbf{99.35}           & \multicolumn{1}{c|}{99.12}             & 96.80                     & 97.93                     & \multicolumn{1}{c|}{\textbf{99.32}}     \\
\multicolumn{1}{|c|}{\textbf{9-NN}}  & \textbf{95.47}        & 83.10                            & \multicolumn{1}{l|}{82.19}        & \textbf{95.90}            & 80.05                    & \multicolumn{1}{c|}{85.06}              & \textbf{88.47}             & 82.42                      & \multicolumn{1}{c|}{82.28}                & 94.52                     & \textbf{99.35}           & \multicolumn{1}{c|}{99.12}             & 96.58                     & 97.93                     & \multicolumn{1}{c|}{\textbf{99.15}}     \\
\multicolumn{1}{|c|}{\textbf{11-NN}} & \textbf{95.16}        & 82.43                            & \multicolumn{1}{l|}{81.53}        & \textbf{95.90}            & 79.11                    & \multicolumn{1}{c|}{83.81}              & \textbf{88.54}             & 82.42                      & \multicolumn{1}{c|}{82.28}                & 94.24                     & \textbf{99.35}           & \multicolumn{1}{c|}{99.12}             & 96.69                     & 97.93                     & \multicolumn{1}{c|}{\textbf{99.15}}     \\
\multicolumn{1}{|c|}{\textbf{13-NN}} & \textbf{95.39}        & 83.10                            & \multicolumn{1}{l|}{82.19}        & \textbf{95.90}            & 79.71                    & \multicolumn{1}{c|}{84.20}              & \textbf{88.25}             & 82.08                      & \multicolumn{1}{c|}{82.94}                & 94.14                     & \textbf{99.35}           & \multicolumn{1}{c|}{99.12}             & 96.48                     & 97.93                     & \multicolumn{1}{c|}{\textbf{99.15}}     \\
\multicolumn{1}{|c|}{\textbf{SVM}}   & \textbf{86.78}        & 80.72                            & \multicolumn{1}{l|}{85.18}        & \textbf{90.00}            & 83.63                    & \multicolumn{1}{c|}{89.46}              & 88.54                      & 84.43                      & \multicolumn{1}{c|}{\textbf{90.55}}       & 93.56                     & \textbf{99.22}           & \multicolumn{1}{c|}{98.77}             & 96.26                     & \textbf{99.64}            & \multicolumn{1}{c|}{99.32}              \\
\multicolumn{1}{|c|}{\textbf{LDA}}   & \textbf{86.47}        & 80.91                            & \multicolumn{1}{l|}{83.96}        & 88.58                     & 82.69                    & \multicolumn{1}{c|}{\textbf{88.70}}     & 87.80                      & 84.32                      & \multicolumn{1}{c|}{\textbf{89.76}}       & 93.18                     & 98.82                    & \multicolumn{1}{c|}{\textbf{99.30}}    & 96.69                     & \textbf{99.03}            & \multicolumn{1}{c|}{98.82}              \\ \hline
\end{tabular}

  \label{tab:accuracyCELEBAvsSINGLEDEEPNETWORK_}
\end{table*}

Experiments were carried out considering images created by STARGAN~\cite{choi2018stargan}, ATTGAN~\cite{he2019attgan}, GDWCT~\cite{cho2019image}, STYLEGAN~\cite{karras2019style}, STYLEGAN2~\cite{karras2020analyzing} and FACEFORENSICS++~\cite{rossler2019faceforensics++} for Deepfake of faces in conjunction with other four Deepfake architectures not dealing with faces: CYCLEGAN~\cite{zhu2017unpaired}, PROGAN~\cite{karras2017progressive}, IMLE~\cite{li2019diverse} and SPADE~\cite{park2019semantic}. Figure~\ref{fig:NetworksDetail} shows a brief presentation of the employed images, the techniques, targets, semantics, etc. by reporting also details about training and testing purposes.

STYLEGAN images~\footnote{\href{https://drive.google.com/drive/folders/1uka3a1noXHAydRPRbknqwKVGODvnmUBX}{https://drive.google.com/drive/folders/STYLEGAN}} and STYLEGAN2 images~\footnote{\href{https://drive.google.com/drive/folders/1QHc-yF5C3DChRwSdZKcx1w6K8JvSxQi7}{https://drive.google.com/drive/folders/STYLEGAN2}} were downloaded from the official websites, while, for images of the other architectures, the pre-trained models were employed to generate them. 
The CT was extracted from all the images with kernels of increasing sizes ($ 3, 5$ and $7$). The CT obtained was employed as input feature vector for different naive classifiers (K-NN, SVM, LDA) with different tasks: (i) discriminating an authentic image from one generated by a specific GAN and (ii) discriminating authentic images from Deepfakes (binary classification - Real Vs Deepfake images generated by all the 10 techniques). We achieved the best classification solution by employing Random Forest as a final binary classifier, thus finalizing the pipeline (Figure~\ref{fig:Pipeline}).

\begin{figure*}[t!]
\begin{center}
\includegraphics[width=0.75\linewidth]{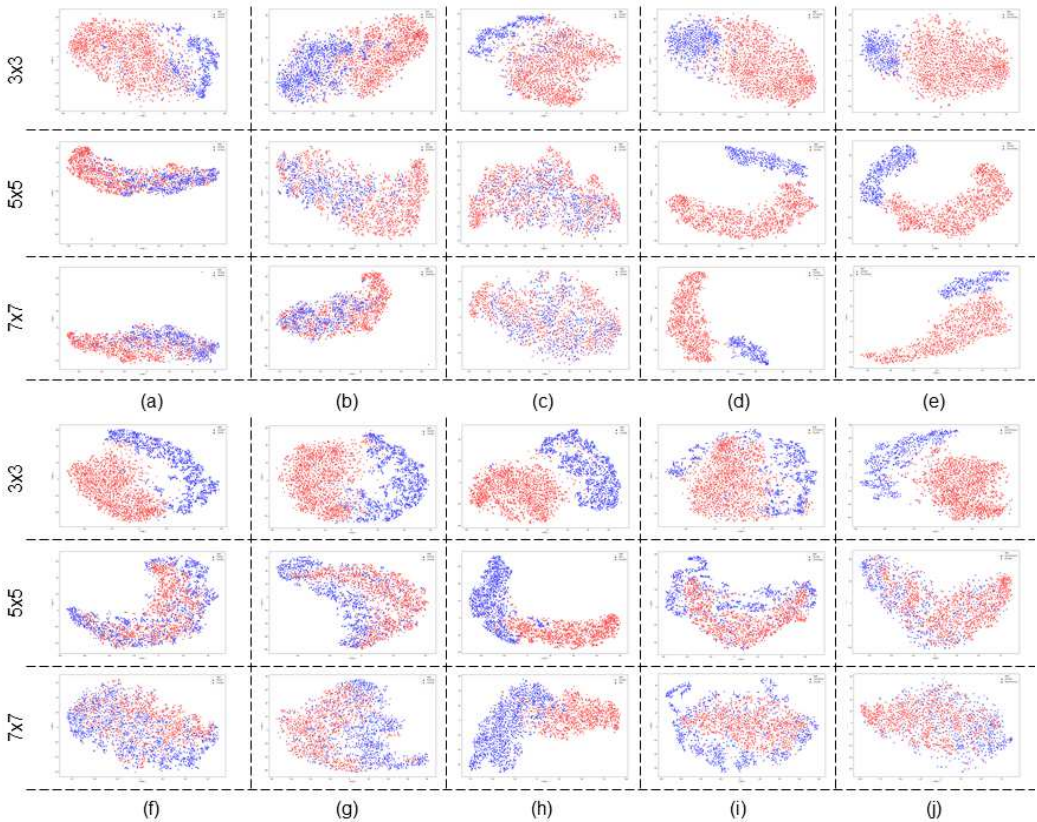}
\end{center}
   \caption{Two-dimensional t-SNE representations (CELEBA: red; DeepNetwork: blue) of all kernel sizes for each classification task: (a) CELEBA – ATTGAN; (b) CELEBA – STARGAN; (c) CELEBA – GDWCT; (d) CELEBA – STYLEGAN; (e) CELEBA – STYLEGAN2; (f) CELEBA - SPADE; (g) CELEBA - PROGAN; (h) CELEBA - IMLE; (i) CELEBA - CYCLEGAN; (j) CELEBA - FACEFORENSICS++.}
\label{fig:tsne10All}
\end{figure*}

Let's first analyse the discriminative power of the CT in order to distinguish authentic images from each of the considered GAN. Figure~\ref{fig:tsne10All} shows a visual representation by means of t-SNE~\cite{maaten2008visualizing}: it is possible to notice how Deepfakes can be ``linearly" separable from authentic samples. Moreover, in most cases the separation is utterly clear. Figure~\ref{fig:tsne10All} visually demonstrates the discriminative power of the extracted CT which, if used as feature vector in a classification task, obtains excellent results as expected. All the classification results are reported in Table~\ref{tab:accuracyCELEBAvsSINGLEDEEPNETWORK_}. In particular, it is possible to note that:

\begin{itemize}
    \item \textbf{CELEBA Vs ATTGAN} the maximum classification accuracy of $92.99\%$, was obtained with KNN (with K = 5, 7), and kernel size of $3x3$.
    \item \textbf{CELEBA Vs CYCLEGAN} the maximum classification accuracy of $93.59\%$, was obtained with KNN (with K = 3), and kernel size of $3x3$.
    \item \textbf{CELEBA Vs FACEFORENSICS++} the maximum classification accuracy of $97.31\%$, was obtained with KNN (with K = 3), and kernel size of $3x3$.
    \item \textbf{CELEBA Vs GDWCT}: the maximum classification accuracy of $91.58\%$, was obtained with KNN (with K = 9) and kernel size of $3x3$.
    \item \textbf{CELEBA Vs IMLE}: the maximum classification accuracy of $97.76\%$, was obtained with KNN (with K = 3) and kernel size of $3x3$.
    \item \textbf{CELEBA Vs PROGAN}: the maximum classification accuracy of $95.85\%$, was obtained with KNN (with K = 5) and kernel size of $3x3$.
    \item \textbf{CELEBA Vs SPADE}: the maximum classification accuracy of $96.72\%$, was obtained with KNN (with K = 3) and kernel size of $3x3$.
    \item \textbf{CELEBA Vs STARGAN}: the maximum classification accuracy of $90.55\%$, was obtained with linear SVM, and kernel size of $7x7$.
    \item \textbf{CELEBA Vs STYLEGAN}: the maximum classification accuracy of $99.48\%$, was obtained with KNN - K = 3, and kernel size of $5x5$.
    \item \textbf{CELEBA Vs STYLEGAN2}: the maximum classification accuracy of $99.64\%$, was obtained with linear SVM, and kernel size of $5x5$.
\end{itemize}

This leads to an empirical hypothesis: the kernel size used by output layers in Deepfake generation techniques is related to the kernel size parameter employed by the CT extraction approach. However, it has to be noted that, on average the kernel size of $3x3$ achieves best results among all the classification tests.

Another interesting insight is that the extracted CT is able to discriminate between images from STYLEGAN and STYLEGAN2: a binary test carried out to discriminate between images from the two ``similar" techniques achieved a maximum accuracy of $99.31\%$ (Table~\ref{tab:accuracyST1ST2}). As stated by the authors of the STYLEGAN2 architecture, they have only updated parts of the generator $G$, in order to remove imperfections of the original STYLEGAN. This further confirms the former hypothesis,  since even a slight modification of $G$, leaves different traces in the images generated and the CT is able to extract such fingerprint.

\begin{table}[t]
\centering
\caption{Accuracy values for binary test between STYLEGAN and STYLEGAN2 with different classifiers and kernel sizes ($3x3$, $5x5$, $7x7$).}
\begin{tabular}{c|ccc|}
\cline{2-4}
\multicolumn{1}{l|}{}                     & \multicolumn{3}{c|}{\begin{tabular}[c]{@{}c@{}}\textbf{STYLEGAN} \\ \textbf{Vs}\\ \textbf{STYLEGAN2}\end{tabular}} \\
                                    \cline{2-4} 
                           & \multicolumn{3}{c|}{\textbf{Kernel Size}}                     \\
                           & \textbf{3x3}  & \textbf{5x5}   & \textbf{7x7}   \\ \hline
\multicolumn{1}{|c|}{\textbf{3-NN}}  & 89.36                & \textbf{90.51} & 87.24          \\
\multicolumn{1}{|c|}{\textbf{5-NN}}  & 89.56               & \textbf{89.87} & 85.52          \\
\multicolumn{1}{|c|}{\textbf{7-NN}}  & 89.16                & \textbf{90.93} & 87.59          \\
\multicolumn{1}{|c|}{\textbf{9-NN}}  & 88.55                & \textbf{89.87} & 87.93          \\
\multicolumn{1}{|c|}{\textbf{11-NN}} & 88.35               & \textbf{90.30} & 87.24          \\
\multicolumn{1}{|c|}{\textbf{13-NN}} & 89.36              & \textbf{89.66} & 87.93          \\
\multicolumn{1}{|c|}{\textbf{SVM}}   & 91.77             & 99.16          & \textbf{99.31} \\
\multicolumn{1}{|c|}{\textbf{LDA}}   & 91.16               & \textbf{98.73} & 98.28          \\ \hline
\end{tabular}

  \label{tab:accuracyST1ST2}
\end{table}

\begin{table}[t]
\centering
\caption{Accuracy values obtained in the binary classification task between Real images vs. images generated by 10 Deepfake architectures. Results are reported with different kernel sizes ($3x3$,, $5x5$, $7x7$) and classifiers trained on 70\% of the dataset and tested on the remaining part. Results are the average accuracy value obtained on a 5-fold cross validation test.}
\begin{tabular}{c|ccc|}
\cline{2-4}
\multicolumn{1}{l|}{}                     & \multicolumn{3}{c|}{\begin{tabular}[c]{@{}c@{}}\textbf{CELEBA} \\ \textbf{Vs}\\ \textbf{DeepNetworks}\end{tabular}} \\ \cline{2-4} 
\multicolumn{1}{l|}{}                     & \multicolumn{3}{c|}{\textbf{Kernel Size}}                                                         \\
\multicolumn{1}{l|}{}                     & \textbf{3x3}                    & \textbf{5x5}                  & \textbf{7x7}                    \\ \hline
\multicolumn{1}{|c|}{\textbf{3-NN}}       & \textbf{89.80}                  & 77.38                         & 78.63                           \\
\multicolumn{1}{|c|}{\textbf{5-NN}}       & \textbf{90.79}                  & 77.20                         & 77.80                           \\
\multicolumn{1}{|c|}{\textbf{7-NN}}       & \textbf{90.44}                  & 76.47                         & 78.39                           \\
\multicolumn{1}{|c|}{\textbf{9-NN}}       & \textbf{90.30}                  & 77.20                         & 78.28                           \\
\multicolumn{1}{|c|}{\textbf{11-NN}}      & \textbf{89.80}                  & 77.29                         & 77.45                           \\
\multicolumn{1}{|c|}{\textbf{13-NN}}      & \textbf{89.73}                  & 77.66                         & 77.69                           \\
\multicolumn{1}{|c|}{\textbf{SVMLinear}}  & \textbf{84.14}                  & 76.28                         & 80.28                           \\
\multicolumn{1}{|c|}{\textbf{SVMSigmoid}} & 58.57                           & 61.36                         & \textbf{63.52}                  \\
\multicolumn{1}{|c|}{\textbf{SVMrbf}}     & \textbf{91.22}                  & 80.04                         & 80.87                           \\
\multicolumn{1}{|c|}{\textbf{SVMPoly}}    & \textbf{88.74}                  & 78.66                         & 78.87                           \\
\multicolumn{1}{|c|}{\textbf{LDA}}        & \textbf{83.50}                  & 77.38                         & 78.98                           \\ \hline
\multicolumn{1}{|c|}{\textbf{Random Forest}}        &   \textbf{98.07}           &          93.81           &        91.22                  \\ \hline
\end{tabular}

  \label{tab:accuracyCELEBAvsALLDEEPNETWORK}
\end{table}

\begin{figure*}[t!]
\begin{center}
\includegraphics[width=0.7\linewidth]{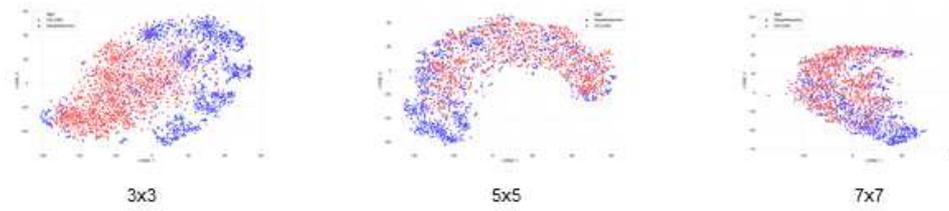}
\end{center}
   \caption{Two-dimensional t-SNE representation (CELEBA: red; All 10 DeepNetworks: blue) of a binary classification problem (with different kernel size): CELEBA Vs All 10 DeepNetworks.}
\label{fig:tsne10Binary}
\end{figure*}

\begin{figure*}[t]
\centering
     \hfill
     \subfloat[\label{subfig:Corrette}]{%
       \includegraphics[width=0.3\linewidth]{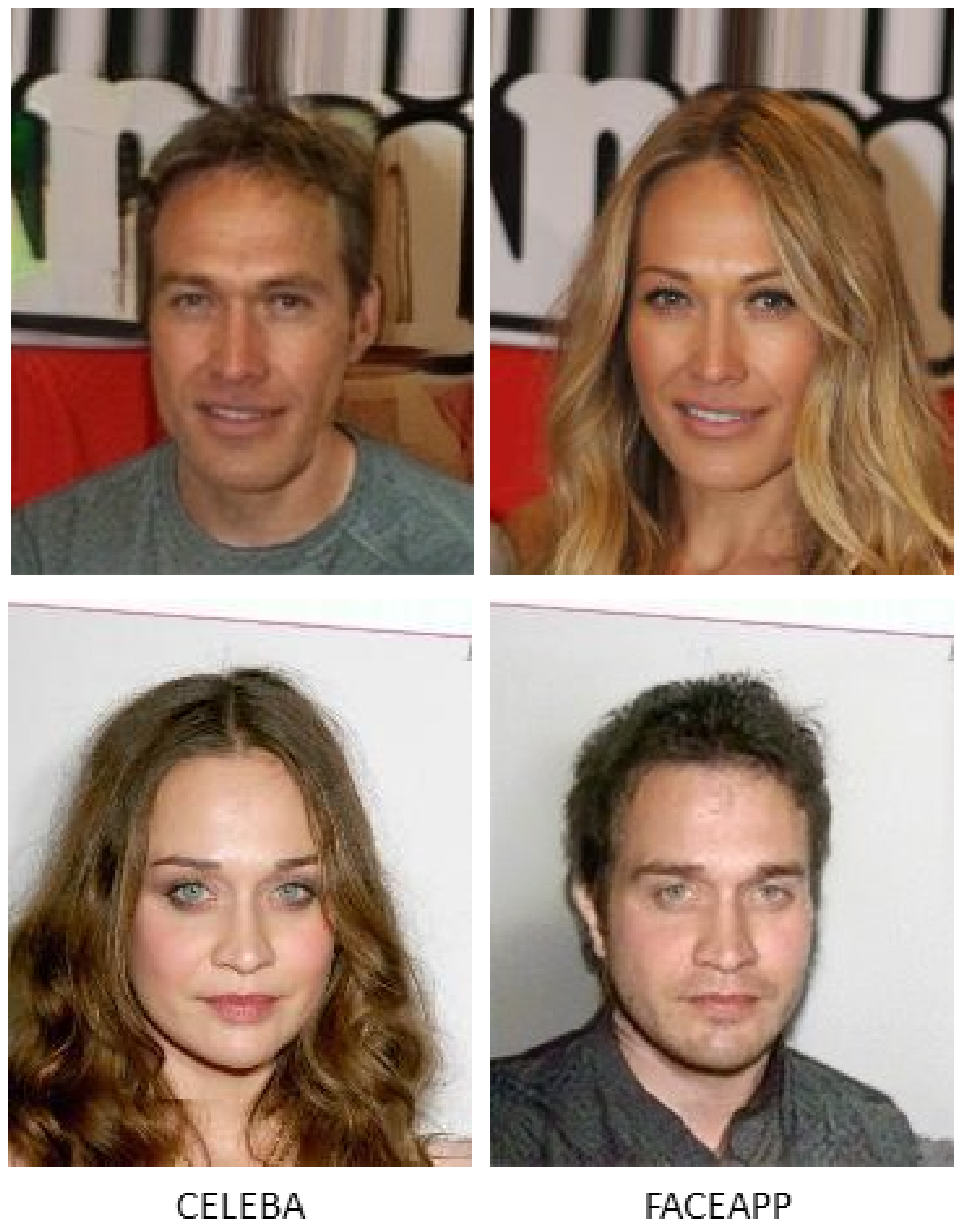}
     }
     \hfill
     \subfloat[\label{subfig:Sbagliate}]{%
       \includegraphics[width=0.3\linewidth]{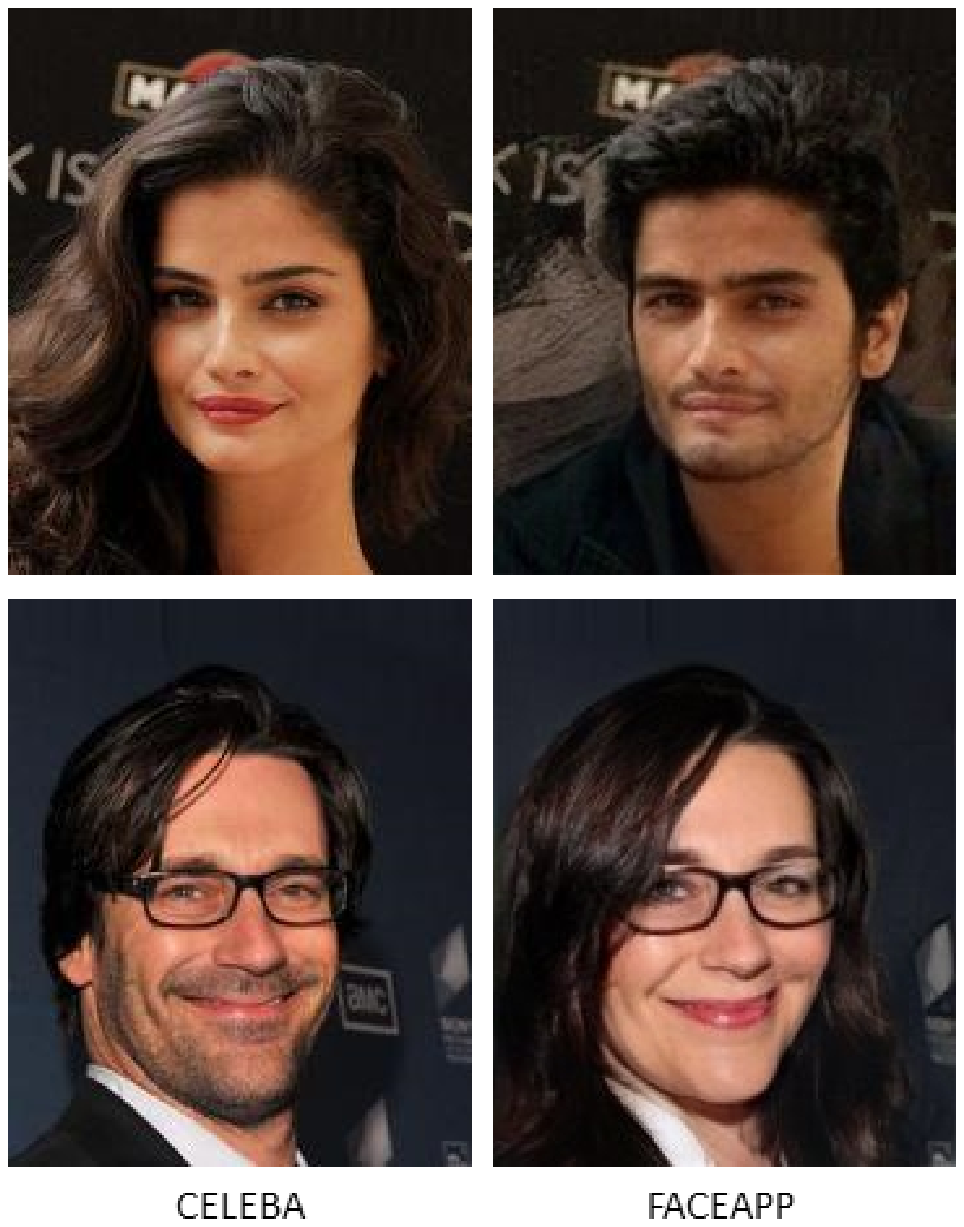}
     }
    \hfill
    \subfloat{}
     \caption{(a) Example of images generated by FACEAPP with correct classification (deepfake) (b) Example of images generated by FACEAPP with incorrect classification (real).}
     \label{fig:FaceApp}
\end{figure*}

We also employed binary classification between real images and Deepfakes coming from all the 10 architectures taken into account. At first, another t-SNE representation was built in order to understand sample separability and distribution in two-dimensional plane. Figure~\ref{fig:tsne10Binary} shows that, in this case, samples cannot be linearly separated thus we carried out tests looking for non-linear classifiers. Indeed, final results demonstrated and confirmed such insights. Best accuracy score was obtained by employing Random Forest properly
chosen as the final step of the Deepfake detection pipeline (Figure \ref{fig:Pipeline}) with a solid 98\% of accuracy (Table~\ref{tab:accuracyCELEBAvsALLDEEPNETWORK}) obtained in our tests.

In this Section, experimental results and t-SNE visualizations demonstrated the discriminative power of the CT extracted from Deepfakes. Moreover, the CT achieves good results in detecting Deepfakes not representing faces, hence demonstrating CT being independent to semantics. To further evaluate the proposed pipeline we employed an additional classification test: detecting Deepfakes created by the famous mobile app FACEAPP.

Recently, the mobile application called FACEAPP is having a lot of success due to the ability to change features of the input image of a face such as gender, age, hair style, etc. The images thus produced are utterly realistic. Hence, a test for automatic detection of Deepfakes produced by FACEAPP has been carried out employing the CT extraction method and the Random Forest classifier already trained for the test previously described. No further training was done on FACEAPP images. For experiments a dataset of Deepfake images was created starting from CELEBA images by using the Android version of FACEAPP (we employed the paid version that does not introduce watermarks on images): 471 images were generated with FACEAPP by applying gender swap on original images. CTs were extracted with kernel size $3x3$ and employed as input for the pre-trained Random Forest classifier. Among the 471 images, 437 were correctly classified as Deepfakes while 34 images were classified as real faces. Figure~\ref{subfig:Corrette} shows two examples of correct classifications while Figure~\ref{subfig:Sbagliate} shows two examples of misclassification. It has to be noted that incorrect classifications are probably due to low light conditions or too few changes in the original images thus making difficult to extract a discriminative CT. According to the reported results we proved the effectiveness of the proposed Deepfake detection technique in a real-case scenario.

\section{Robustness Experiments}
\label{sec:RobustTest}

\begin{figure*}[t!]
\begin{center}
\includegraphics[width=0.9\linewidth]{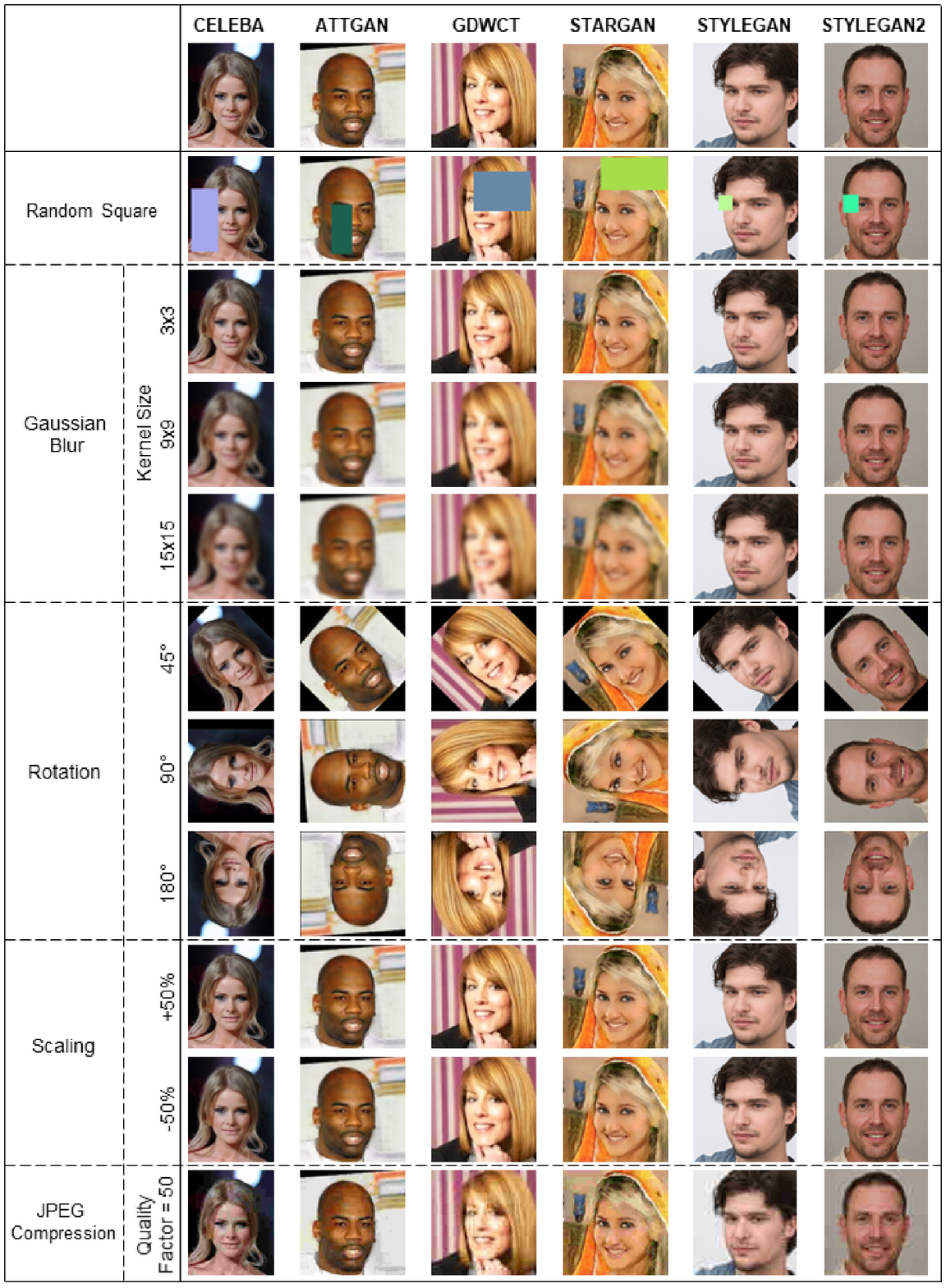}
\end{center}
   \caption{Examples of real (CELEBA) and deepfake images of faces (ATTGAN, GDWCT, STARGAN, STYLEGAN, STYLEGAN2) with six different kind of attacks: \textit{Random Square}, \textit{Gaussian Blur}, \textit{Rotation}, \textit{Scaling} and \textit{JPEG Compression}.}
\label{fig:Filters}
\end{figure*}

\begin{table*}[t!]
\centering
\caption{Robustness to attacks: table reports accuracy values obtained (percentage) for the binary classification task (real vs. Deepfakes) employing the final classification solution for each different kernel size (3x3, 5x5. 7x7). The final classifier was trained on the augmented dataset (70\% of data for the training set) and 5-fold cross-validated. The first row represents the accuracy obtained by the trained \emph{robust} classifier without any attack.}
  \begin{adjustbox}{max width=\textwidth}
\begin{tabular}{c|ccc|ccc|ccc|ccc|ccc|}
\cline{2-16}
\multicolumn{1}{l|}{}                                                                                       & \multicolumn{3}{c|}{\textbf{ATTGAN}}       & \multicolumn{3}{c|}{\textbf{GDWCT}}        & \multicolumn{3}{c|}{\textbf{STARGAN}}      & \multicolumn{3}{c|}{\textbf{STYLEGAN}}     & \multicolumn{3}{c|}{\textbf{STYLEGAN2}}    \\ \cline{2-16} 
\multicolumn{1}{l|}{}                                                                                       & \multicolumn{3}{c|}{\textbf{Kernel Size}}  & \multicolumn{3}{c|}{\textbf{Kernel Size}}  & \multicolumn{3}{c|}{\textbf{Kernel Size}}  & \multicolumn{3}{c|}{\textbf{Kernel Size}}  & \multicolumn{3}{c|}{\textbf{Kernel Size}}  \\
\multicolumn{1}{l|}{}                                                                                       & \textbf{3x3} & \textbf{5x5} & \textbf{7x7} & \textbf{3x3} & \textbf{5x5} & \textbf{7x7} & \textbf{3x3} & \textbf{5x5} & \textbf{7x7} & \textbf{3x3} & \textbf{5x5} & \textbf{7x7} & \textbf{3x3} & \textbf{5x5} & \textbf{7x7} \\ \hline

\multicolumn{1}{|c|}{\begin{tabular}[c]{@{}c@{}}\textbf{Raw} \\ \textbf{Images}\end{tabular}}                     & 92.99        & 88.51        & 87.11           & 91.58        & 77.41       & 82.01        & 88.54        & 84.43        & 90.55        & 95.29        & 99.48          & 99.30        & 96.91       & 99.64          & 99.32          \\ \hline

\multicolumn{1}{|c|}{\begin{tabular}[c]{@{}c@{}}\textbf{Random} \\ \textbf{Square}\end{tabular}}                     & 82.54        & 75.47        & 75           & 62.03        & 61.54        & 63.27        & 81.16        & 78.95        & 76.19        & 97.26        & 100          & 97.37        & 99.02        & 100          & 100          \\ \hline
\multicolumn{1}{|c|}{\begin{tabular}[c]{@{}c@{}}\textbf{Gaussian Blur}.\\ \textbf{kernel size = 3x3}\end{tabular}}   & 77.78        & 73.58        & 72.22        & 56.96        & 59.38        & 61.22        & 73.91        & 80.7         & 61.9         & 93.15        & 98.33        & 92.11        & 96.08        & 98.81        & 96.08        \\ \hline
\multicolumn{1}{|c|}{\begin{tabular}[c]{@{}c@{}}\textbf{Gaussian Blur.}\\ \textbf{kernel size = 9x9}\end{tabular}}   & 76.19        & 76.92        & 68.57        & 56.96        & 67.19        & 61.22        & 72.46        & 77.19        & 64.29        & 97.26        & 100          & 94.59        & 96.08        & 97.62        & 94.12        \\ \hline
\multicolumn{1}{|c|}{\begin{tabular}[c]{@{}c@{}}\textbf{Gaussian Blur.}\\ \textbf{kernel size = 15x15}\end{tabular}} & 80.95        & 76.92        & 77.14        & 64.56        & 67.69        & 57.14        & 82.61        & 80.7         & 75.61        & 97.26        & 98.33        & 94.59        & 100          & 97.59        & 98.04        \\ \hline
\multicolumn{1}{|c|}{\begin{tabular}[c]{@{}c@{}}\textbf{Rotation} \\ \textbf{45°}\end{tabular}}                      & 90           & 84.31        & 85.29        & 67.53        & 73.02        & 66.67        & 85.29        & 82.14        & 87.8         & 89.04        & 91.67        & 91.89        & 97.4         & 94.2         & 97.62        \\ \hline
\multicolumn{1}{|c|}{\begin{tabular}[c]{@{}c@{}}\textbf{Rotation} \\ \textbf{90°}\end{tabular}}                      & 100          & 94.23        & 100          & 93.59        & 92.19        & 93.75        & 92.75        & 92.98        & 97.56        & 100          & 100          & 97.3         & 100          & 100          & 100          \\ \hline
\multicolumn{1}{|c|}{\begin{tabular}[c]{@{}c@{}}\textbf{Rotation} \\ \textbf{180°}\end{tabular}}                     & 83.87        & 86.54        & 82.86        & 74.36        & 67.19        & 59.18        & 84.06        & 91.23        & 78.57        & 100          & 100          & 91.89        & 97.03        & 98.8         & 98.04        \\ \hline
\multicolumn{1}{|c|}{\begin{tabular}[c]{@{}c@{}}\textbf{Scaling} \\ \textbf{+50\%}\end{tabular}}                     & 88.71        & 78.43        & 91.18        & 78.21        & 71.88        & 68.09        & 89.71        & 83.93        & 90           & 97.22        & 100          & 97.3         & 99           & 98.78        & 100          \\ \hline
\multicolumn{1}{|c|}{\begin{tabular}[c]{@{}c@{}}\textbf{Scaling} \\ \textbf{-50\%}\end{tabular}}                     & 75.81        & 78.85        & 77.78        & 71.79        & 57.81        & 68.09        & 79.71        & 64.91        & 64.29        & 95.83        & 96.67        & 100          & 99.01        & 97.59        & 94.23        \\ \hline
\multicolumn{1}{|c|}{\begin{tabular}[c]{@{}c@{}}\textbf{JPEG} \\ \textbf{Compression}\end{tabular}}                  & 86.69        & 91.67        & 91.18        & 85.17        & 89.33        & 84.66        & 89.17        & 92.69        & 92.01        & 99.5         & 99.33        & 97.57        & 99.49        & 98.96        & 98.55        \\ \hline
\end{tabular}
\end{adjustbox}

  \label{tab:RobustnessTest}
\end{table*}

Finally, we introduce further tests about overall robustness. A series of attacks were made at different Deepfake images of faces generated by ATTGAN, GDWCT, STARGAN, STYLEGAN and STYLEGAN2 and real images (CELEBA). In particular, the following attacks were carried out:

\begin{enumerate}
    \item \textit{Adding one rectangle with different sizes, positions and colors at random}: in this way details are removed. Since the CT extracts information from pixel correlations, the addition of this rectangle could lead to errors. This could happen specifically for STARGAN or ATTGAN considering that they change only few elements in a face (e.g. hair color) and if these elements are removed by the rectangle low classification accuracy values are expected;
    \item \textit{Adding Gaussian Blur with different kernel sizes (3x3, 9x9, 15x15)}: the noise added to the images could destroy the pixels correlation created by Deepfake architectures and remove the CT;
    \item \textit{Rotating images by 45, 90, 180 degrees}: rotations could lead to interpolation transformation with modification on CTs similar to the Gaussian blur attack;
    \item \textit{Scaling images (+50\%, -50\%)}: due to the interpolation operations carried out, information will be added or removed. CT extracted from images with high details (such as those of STYLEGAN and STYLEGAN2) would be more robust to this type of operation;
    \item \textit{JPEG compression with quality factor equal to 50}: in general, a compression operation (such as JPEG) removes high frequency information which could be of major importance for the CT discriminative power. Moreover, a JPEG compression with Quality Factor 50 is similar to those applied by social networks such as Facebook or Instant Messengers like Whatsapp~\cite{giudice2017classification}, making this test another real-case scenario.
\end{enumerate}

\begin{figure*}[t!]
\begin{center}
\includegraphics[width=0.75\linewidth]{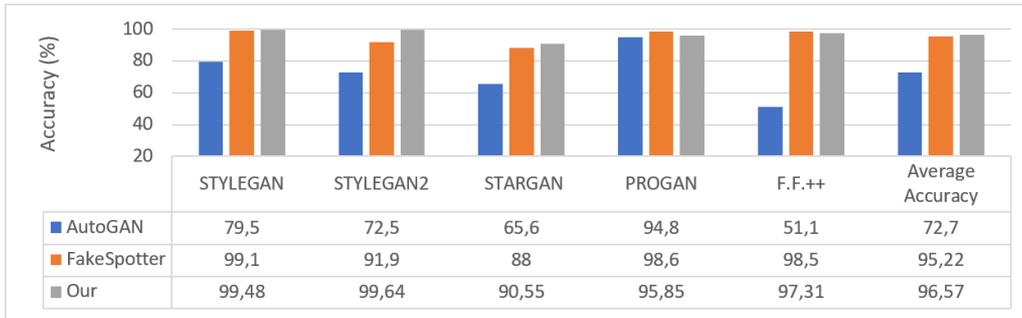}
\end{center}
   \caption{Comparison of the proposed approach (Our) vs.  FakeSpotter~\cite{wang2019fakespotter}  and AutoGAN~\cite{zhang2019detecting}.}
\label{fig:FakeSpotter}
\end{figure*}

\begin{figure*}[t!]
\begin{center}
\includegraphics[width=0.75\linewidth]{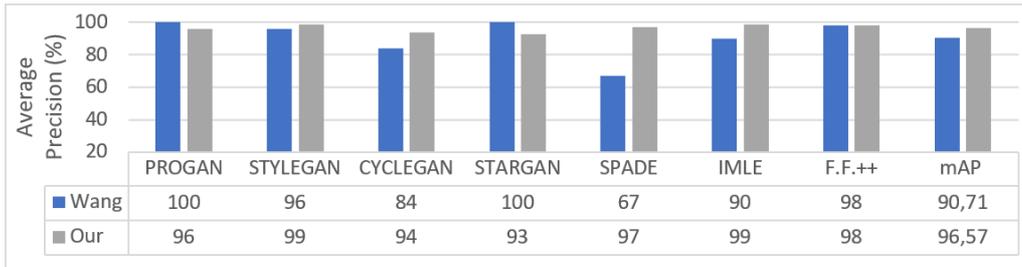}
\end{center}
   \caption{Comparison of the proposed approach (Our) vs. Wang et al.~\cite{Wang_2020_CVPR}.}
\label{fig:Wang}
\end{figure*}

Once the above mentioned filters are applied individually to images, the CT extraction method was applied and Real Vs Deepfake classifications carried out against each GAN (e.g. CELEBA\textsubscript{RandomSquare} Vs GDWCT\textsubscript{RandomSquare}, CELEBA\textsubscript{GaussianBlur} Vs STARGAN\textsubscript{GaussianBlur}, etc.). 

The classification results are reported in Table~\ref{tab:RobustnessTest}~\footnote{We report in this table the maximum accuracy classification value obtained through k-NN (with $k=\{3,5,7,9,11,13\}$), LDA (Linear Discriminant Analysis), SVM (Support Vector Machine) with linear kernel and Random Forest}. 

Figure~\ref{fig:Filters} shows an example of images obtained after operations listed before. It is possible to observe that the dataset plays a fundamental role: the output of ATTGAN, STARGAN and GDWCT and the output of STYLEGAN and STYLEGAN2 after the Gaussian Blur operation: images from ATTGAN, STARGAN and GDWC show a greater visible blur (and therefore a worse visual quality with greater lack of details) respect to STYLEGAN and STYLEGAN2 images. This is mainly determined by the capability of STYLEGAN and STYLEGAN2 to create images of a bigger size. 

Results reported in Table~\ref{tab:RobustnessTest} show that the CT extracted is robust to almost all considered attacks. 

In particular, as stated before, STYLEGAN and STYLEGAN2 images obtained the best classification accuracy values (Real Vs Deepfake) due to their bigger original size.  GDWCT, which creates the smallest images (Figure~\ref{fig:NetworksDetail}), is the least robust to attacks and maintains a proper accuracy result comparable with results without attacks only for JPEG compression. 

However, another interesting insight comes from the rotation attacks: a rotation of 90 degrees anticlockwise, which is a rotation that does not introduce interpolation, unexpectedly produces better classification results for each of the considered Deepfake architecture. This could be related to a specific \emph{major} direction of the CT and should be better investigated in future works.

\section{Comparisons with Deepfake detection methods}
\label{sec:CompDetMeth}

Section \ref{sec:related} presented a detailed discussion of the state-of-the-art in the field of Deepfakes and specifically in Section \ref{sec:detection_methods} the detection methods available as today were discussed.

While analytical techniques based on frequency domain still lack of accuracy, CNN based techniques seems to achieve good results but tend to be context-dependent, prone to overfitting and provably depending to high-level semantics extracted from images. Moreover, CNN techniques are computationally intensive and difficult to be explained or controlled. 
In  \cite{hulzebosch2020detecting} the authors discussed this limit about CNN. We carried out tests with a deep neural network  VGG-16\footnote{\url{https://github.com/1297rohit/VGG16-In-Keras}} - on spatial and frequency domains - to solve the binary classification task (Real Vs All 10 Deepfakes) on the datasets described above, obtaining the best result equal to only 53\% of accuracy (similar to the random classifier). Better results are achievable by only a complex deep neural network architecture, is what done by recent state-of-the-art methods.
The more the architecture is complex and the more is not only an higher computational power is necessary but it is more complicated understand what are the high level features that the network used (probably these features are focused on differences in eye sizes, a-symmetries, etc.) to distinguish Real Vs Deepfakes images.

In this Section a detailed discussion is carried out, comparing results obtained by the proposed approach with the best literature methods: Wang et al.~\cite{wang2019fakespotter} (the authors of FakeSpotter), Zhang et al.~\cite{zhang2019detecting} (the authors of AutoGAN) and Wang et al. \cite{Wang_2020_CVPR} were taken into account for comparisons in the Real Vs. Deepfake binary classification task.

For FakeSpotter and AutoGAN,  Deepfakes from STYLEGAN, STYLEGAN2, STARGAN, PROGAN and FACEFORENSICS++ architectures were taken into account. Results of this comparison are reported in Figure~\ref{fig:FakeSpotter}. It is possibile to note that, not only we obtained accuracy values of over 90\% in all cases, but we overcame FakeSpotter on the average accuracy evaluation. Only in the case of PROGAN and FACEFORENSICS++ we obtained a slightly lower value.

In Wang et al.~\cite{Wang_2020_CVPR}, the following seven Deepfake architectures were taken into account for a fair comparison: PROGAN, STYLEGAN, CYCLEGAN, STARGAN, SPADE, IMLE, FACEFORENSICS++. Wang et al. reported results in the binary classification task as Average Precision between different datasets: images with no data augmentation; images with Gaussian blur added; images JPEG compressed; images both blurred and JPEG compressed. Figure~\ref{fig:Wang} shows the comparison results obtained by Wang et al. and the proposed approach, reporting the Average Precision (AP) and mean Average Precision (mPA) values for each different Deepfake architecture. It is possible to note that the proposed method obtains better results specifically on Deepfakes of SPADE, IMLE and CYCLEGAN: architectures that do not produce images of faces, furtherly demonstrating the robustness of the extracted CT and classification pipeline to semantics of the image.

\section{Conclusions and future works}
\label{sec:conclusion}

In this paper, a finalization of a former work on analysis of Deepfake images was presented. An algorithm based on Expectation-Maximization was employed to extract the Convolutional Traces (CT): a sort of unique fingerprint useable to identify not only if an image is a Deepfake but also the GAN architecture that generated it. The CT extracted is a fingerprint demonstrated to have high discriminative power, robustness to attacks and independence to high-level concepts of images (semantics). Obtained results demonstrate also to overcome the state-of-the-art with a technique simple and fast to be computed. Indeed the CT is related to the generation process of images and further better results can be obtained by rotating input images in order to find the \emph{most important} direction. This particular hint will be investigated in future works.

\section*{Acknowledgement}
This research was supported by iCTLab s.r.l. - Spin-off of University of Catania (\url{https://www.ictlab.srl}), which provided domain expertise and computational power that greatly assisted the activity.

\label{sect:bib}
\balance
\bibliographystyle{IEEEtran}
\bibliography{main}

\end{document}